%% file: main.tex
\documentclass[10pt,twocolumn,letterpaper]{article}

\usepackage[pagenumbers]{iccv} 

\input{preamble}
\definecolor{iccvblue}{rgb}{0.21,0.49,0.74}
\usepackage[pagebackref,breaklinks,colorlinks,allcolors=iccvblue]{hyperref}

\title{DidSee: Diffusion-Based Depth Completion for Material-Agnostic Robotic Perception and Manipulation}

\author{
Wenzhou Lyu$^{1}$\quad Jialing Lin$^{1}$\quad Wenqi Ren$^{1}$\quad Ruihao Xia$^{1}$\quad Feng Qian$^{1}$ \quad Yang Tang$^{1}$
\\[2mm]
$^1$East China University of Science and Technology
\\[2mm]
}

\begin{document}
\twocolumn[{%
\renewcommand\twocolumn[1][]{#1}%
\maketitle
\begin{center}
\centering
\captionsetup{hypcap=false}
\vspace{-2em}
\includegraphics[width=\linewidth]{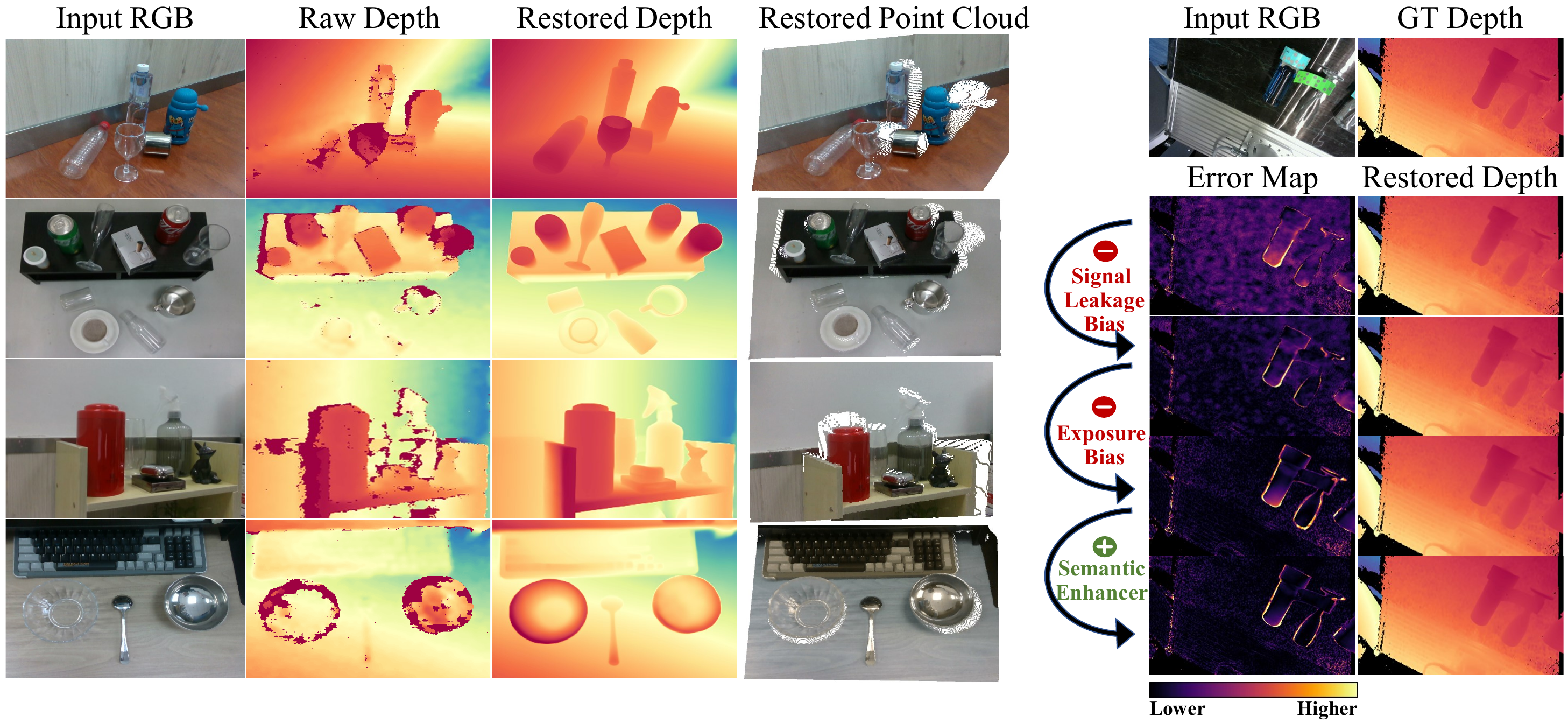}
\vspace{-1.5em}
\captionof{figure}{\textbf{We propose DidSee, a diffusion-based depth completion framework for non-Lambertian objects.} DidSee reduces depth restoration errors by mitigating \textit{signal leakage} and \textit{exposure biases} while incorporating a novel \textit{semantic enhancer} (Right). Without scaling up data size, it generalizes robustly to in-the-wild scenes (Left) and facilitates material-agnostic robotic perception and manipulation in real-world scenarios.}
\vspace{-5pt}
\label{fig:teaser}
\end{center}
}]

\input{sec/0_abstract}
\input{sec/1_intro}
\input{sec/2_related_work}

\input{sec/3_preliminary}
\input{sec/4_method}
\input{sec/5_exp}
\input{sec/6_conclusion}

\clearpage
{
    \small
    \bibliographystyle{ieeenat_fullname}
    \bibliography{main}
}
\input{sec/7_appendix}
\end{document}

%% file: preamble.tex
\definecolor{first}{rgb}{1.0, .83, 0.3}
\definecolor{second}{rgb}{1.0, 0.93, 0.7}

\def \first {\cellcolor{red!15}\textbf}
\def \second {\cellcolor{blue!15}}

\def\para#1{\vspace{0.25em}\noindent\textbf{#1}}

\usepackage{algpseudocode}
\usepackage{multirow}
\usepackage{newtxmath}
\usepackage{booktabs} 
\usepackage{amsmath}
\usepackage{bbding}
\usepackage{xcolor}
\usepackage{colortbl}
\usepackage{makecell}
\usepackage{pifont}
\usepackage{times}
\usepackage{epsfig}
\usepackage{graphicx}
\usepackage{amsmath}
\usepackage{amssymb}
\usepackage{caption}

%% file: sec/0_abstract.tex
\begin{abstract}
Commercial RGB-D cameras often produce noisy, incomplete depth maps for non-Lambertian objects. Traditional depth completion methods struggle to generalize due to the limited diversity and scale of training data. Recent advances exploit visual priors from pre-trained text-to-image diffusion models to enhance generalization in dense prediction tasks. However, we find that biases arising from training-inference mismatches in the vanilla diffusion framework significantly impair depth completion performance. Additionally, the lack of distinct visual features in non-Lambertian regions further hinders precise prediction. To address these issues, we propose \textbf{DidSee}, a diffusion-based framework for depth completion on non-Lambertian objects. First, we integrate a rescaled noise scheduler enforcing a zero terminal signal-to-noise ratio to eliminate signal leakage bias. Second, we devise a noise-agnostic single-step training formulation to alleviate error accumulation caused by exposure bias and optimize the model with a task-specific loss. Finally, we incorporate a semantic enhancer that enables joint depth completion and semantic segmentation, distinguishing objects from backgrounds and yielding precise, fine-grained depth maps. DidSee achieves state-of-the-art performance on multiple benchmarks, demonstrates robust real-world generalization, and effectively improves downstream tasks such as category-level pose estimation and robotic grasping.
\end{abstract}

%% file: sec/1_intro.tex
\section{Introduction}
Accurate RGB-D data is essential in robotics tasks such as scene understanding~\cite{pan2025discovering,Koch_2024_Open3DSG,Wang_2024_CVPR}, 3D reconstruction~\cite{Wang_2024_MorpheuS,zhang2024omni6dpose,Fu_2025_eccv}, and robotic manipulation~\cite{Ze2024DP3, NEURIPS2024_8e5dc596,huang2024fourier}. However, commercial RGB-D sensors often perform poorly on non-Lambertian surfaces (\eg, transparent or specular materials), generating noisy and incomplete depth maps~\cite{10288041}. To address this limitation, recent approaches leverage RGB images to guide the recovery of accurate depth from partial measurements~\cite{ClearGrasp,LIDF,DFNet,FDCT,TODE-Trans,SwinDR,TranspareNet,TCRNet}. Although these methods have achieved notable improvements, accurately predicting depth for non-Lambertian objects remains challenging due to inherent ambiguities in visual features. In particular, foreground-background color blending can confound correspondence estimation~\cite{10288041,Wen_2024_layeredflow}. Moreover, these methods struggle to generalize to real-world scenarios containing novel objects and complex spatial relationships, due to the limited diversity and scale of training data~\cite{DFNet}.

Beyond scaling data size, recent studies have explored diffusion priors for generalizable dense prediction~\cite{Marigold, Geowizard, E2EFT, Genpercept, Lotus, D3roma}. These works demonstrate that text-to-image diffusion models, such as Stable Diffusion~\cite{Stable-Diffusion}, pre-trained on internet-scale image collections, possess powerful and comprehensive visual priors that can enhance dense prediction performance. However, most of these methods directly adapt Stable Diffusion~\cite{Stable-Diffusion} for dense prediction tasks, ignoring the negative impact caused by disparities between generation and dense prediction. For example, the \textit{signal leakage bias}~\cite{Everaert_2024_WACV,Lin_2024_WACV} and \textit{exposure bias}~\cite{Everaert_2024_WACV,Li_2024_ICLR,Li_2024_ICLR_2,ning_2024_ICLR}, which have a relatively minor impact on image generation, can significantly degrade dense predictions (see the right image of \cref{fig:teaser}). In dense prediction tasks such as depth completion, accuracy outweighs visual fidelity.

Motivated by these concerns, we first systematically analyze the limitations of directly adopting the vanilla diffusion framework for depth completion. Our analysis yields several important findings: 1) \textit{Signal leakage bias} introduces unexpected errors at the first inference step, adversely affecting the performance of depth completion models. 2) Multi-step inference exacerbates error accumulation due to \textit{exposure bias}, leading to progressively deteriorating predictions. 3) Non-Lambertian surfaces often lack distinct visual features, making it difficult for the model to produce sharp and accurate depth estimates in these regions.

To address these issues, we propose DidSee, a novel diffusion-based depth completion framework with three key innovations: 1) DidSee adopts a rescaled scheduler that enforces a zero-terminal signal-to-noise ratio (SNR) to mitigate signal leakage. 2) Leveraging the zero terminal-SNR, we introduce an efficient noise-agnostic single-step training formulation to alleviate exposure bias while incorporating a task-specific loss to improve accuracy. 3) We integrate a semantic enhancer that enables the model to jointly generate depth and semantic maps, enhancing its ability to distinguish non-Lambertian surfaces from complex backgrounds, resulting in more precise depth predictions.

Extensive experiments on three benchmark datasets~\cite{DFNet, ClearPose, SwinDR} demonstrate that DidSee achieves state-of-the-art (SoTA) performance. Furthermore, we examine the quality of restored depth on category-level pose estimation and robotic manipulation, where objects exhibit diverse material properties. The results highlight the practical value of DidSee in real-world applications. Our contributions are summarized as follows:
\begin{itemize}
    \item We propose DidSee, a diffusion-based framework for depth completion on non-Lambertian objects.
    \item We identify and address two critical biases in vanilla diffusion models that significantly degrade depth completion accuracy.
    \item We introduce a novel semantic enhancer that improves object-background distinction, enabling more precise and reliable depth predictions.
    \item DidSee achieves SoTA performance across multiple benchmarks, substantially enhancing robotic downstream tasks involving various material objects.
\end{itemize}

%% file: sec/2_related_work.tex
\subsection{Depth Completion for Non-lambertian Objects}
Accurately restoring depth information for non-Lambertian objects remains a significant challenge in robotic perception~\cite{10288041}. Traditional depth completion methods leverage RGB images to infer depth from incomplete and noisy observations~\cite{ClearGrasp, LIDF, DFNet, FDCT, TODE-Trans, SwinDR, TranspareNet, TCRNet}. Existing methods have explored various network architectures, including lightweight U-Net-based models~\cite{DFNet, FDCT}, Vision Transformers~\cite{TODE-Trans}, and Swin Transformers~\cite{SwinDR}. While these approaches have shown promising results, they often suffer from poor generalization, particularly on non-Lambertian surfaces, where transparency and reflectivity obscure reliable visual cues, making depth restoration more challenging. To address this, we exploit diffusion priors to enhance generalizability without additional training data.

\subsection{Diffusion Model for Dense Prediction} 
Recent studies have demonstrated the potential of pre-trained Stable Diffusion~\cite{Stable-Diffusion} in generalized dense prediction tasks~\cite{Marigold, Geowizard, E2EFT, Lotus, D3roma}. Methods like Marigold~\cite{Marigold} and D3RoMa~\cite{D3roma} directly fine-tune Stable Diffusion~\cite{Stable-Diffusion} for depth estimation. However, they overlook the differences between image generation and dense prediction tasks, leading to suboptimal performance. To address these, recent works such as Lotus~\cite{Lotus} and E2E-FT~\cite{E2EFT} have proposed modifications to the diffusion framework. Lotus-G~\cite{Lotus} utilizes $x_0$-prediction and integrates a detail preserver to enhance depth estimation while E2E-FT~\cite{E2EFT} adopts trailing timestep selection to mitigate flaws in noise scheduling. Despite these advancements, diffusion-based methods still suffer from inherent biases that introduce unexpected errors. In this work, we reveal and address these biases while introducing a novel semantic enhancer that improves object-background distinction, enabling more accurate and robust depth completion.

%% file: sec/3_preliminary.tex
\input{figs/fig_overview}

\section{Preliminary}

\para{Diffusion Formulation.} 
Depth completion aims to restore accurate depth values $\mathbf{y^d}$ from incomplete depth maps $\mathbf{d}$ and corresponding RGB images $\mathbf{x}$. In this work, we formulate this problem as an image-conditioned generation task based on Stable Diffusion~\cite{Stable-Diffusion}, which executes the diffusion process within the low-dimensional latent space of a \textit{Variational Autoencoder} (VAE) for computational efficiency. The VAE consists of an encoder $\mathcal{E}$ and a decoder $\mathcal{D}$, which learn bidirectional mappings between image space $\mathcal{X}$ and latent space $\mathcal{Z}$, \ie, $\mathcal{E}(\mathbf{x})=\mathbf{z^x}$, $\mathcal{D}(\mathbf{z^x}) \approx \mathbf{x}$. Recent works~\cite{Marigold, Geowizard, Lotus} have also leveraged the VAE to encode depth maps, \ie, $\mathcal{E}(\mathbf{d})=\mathbf{z^d}$, $\mathcal{D}(\mathbf{z^d}) \approx \mathbf{d}$. 

The diffusion process includes a \textit{forward} and a \textit{reverse} process. In the \textit{forward} process, Gaussian noise is progressively added to the clean sample $\mathbf{z^y_0}$ over $t \in \{1, \dots, T\}$ timesteps, producing the corrupted sample $\mathbf{z}_t^\mathbf{y}$ given by:
\vspace{-3pt}\begin{equation}\vspace{-3pt}
	\label{eqn:noisy_sample}
	\mathbf{z}_t^{\mathbf{y}}=\sqrt{\bar{\alpha}_t} \mathbf{z}_{\mathbf{0}}^{\mathbf{y}}+\sqrt{1-\bar{\alpha}_t} \boldsymbol{\epsilon},
\end{equation}
where $\boldsymbol{\epsilon} \sim \mathcal{N}(0,\mathbf{I})$ and $\bar{\alpha}_t:=\prod_{s=1}^t\left(1-\beta_s \right)$. The forward variances $\{\beta_1, \dots, \beta_T\}$ are predefined hyperparameters. In the \textit{reverse} process, a denoising model $f_\theta$ is trained to progressively remove noise from the corrupted sample $\mathbf{z}_t^\mathbf{y}$ and recover the clean sample $\mathbf{z^y_0}$. Adopting the $v$-prediction~\cite{v-prediction} formulation, the target output of the model, $\mathbf{v}_t$, is computed as: 
\vspace{-3pt}\begin{equation}\vspace{-3pt}
	\label{eqn:v_target}
	\mathbf{v}_t = \sqrt{\bar{\alpha}_t} \boldsymbol{\epsilon} + \sqrt{1-\bar{\alpha}_t} \mathbf{z^y_0}.
\end{equation}
The predicted clean sample $\mathbf{\hat{z}^y_0}$ can be obtained by:
\vspace{-3pt}\begin{equation}\vspace{-3pt}
	\label{eqn:v_hat}
	\hat{\mathbf{v}}_t=f_{\theta}([\mathbf{z^x},\mathbf{z^d},\mathbf{z}_t^\mathbf{y}],t),
\end{equation}
\vspace{-3pt}\begin{equation}\vspace{-3pt}
	\label{eqn:clean_sample}
	\mathbf{\hat{z}^y_0} = \sqrt{\bar{\alpha}_t} \mathbf{z}_t^\mathbf{y} + \sqrt{1-\bar{\alpha}_t} \hat{\mathbf{v}}_t,
\end{equation}
where $\hat{\mathbf{v}}_t$ is the prediction of denoising model.

\input{tables/table_noise_scheduler}
\para{Timestep Selection Strategy.} Diffusion models are typically trained over a large number of timesteps (\eg, $T=1000$) but only perform inference over fewer timesteps (\eg, $S=10$). Recent studies~\cite{Lin_2024_WACV, E2EFT} have identified a critical issue with the conventional \textit{leading} timestep selection strategy, which omits the final step $T$ during inference. For example, when $S=5$, the sampling process starts at $t=801$ (see \cref{tab:timestep_selection}), where the model expects the input $\mathbf{z}_{801}^\mathbf{y}$. Instead, the model receives pure Gaussian noise $\boldsymbol{\epsilon} \approx \mathbf{z}_{1000}^\mathbf{y}$. This mismatch between the expected and actual inputs produces inconsistent pairings of timesteps and noise, leading to suboptimal predictions. To resolve this, E2E-FT~\cite{E2EFT} adopts the \textit{trailing} timestep selection strategy~\cite{Lin_2024_WACV}, which ensures the final timestep $T$ is consistently included during inference, as illustrated in \cref{tab:timestep_selection}. Nevertheless, two inherent biases—signal leakage bias and exposure bias—within the diffusion framework degrade prediction accuracy, even when employing the trailing strategy. Moreover, the absence of distinct visual features in non-Lambertian surfaces poses an additional challenge for precise depth completion.

%% file: figs/fig_overview.tex
\begin{figure*}[!htbp]
\centering
\includegraphics[width=0.97\textwidth]{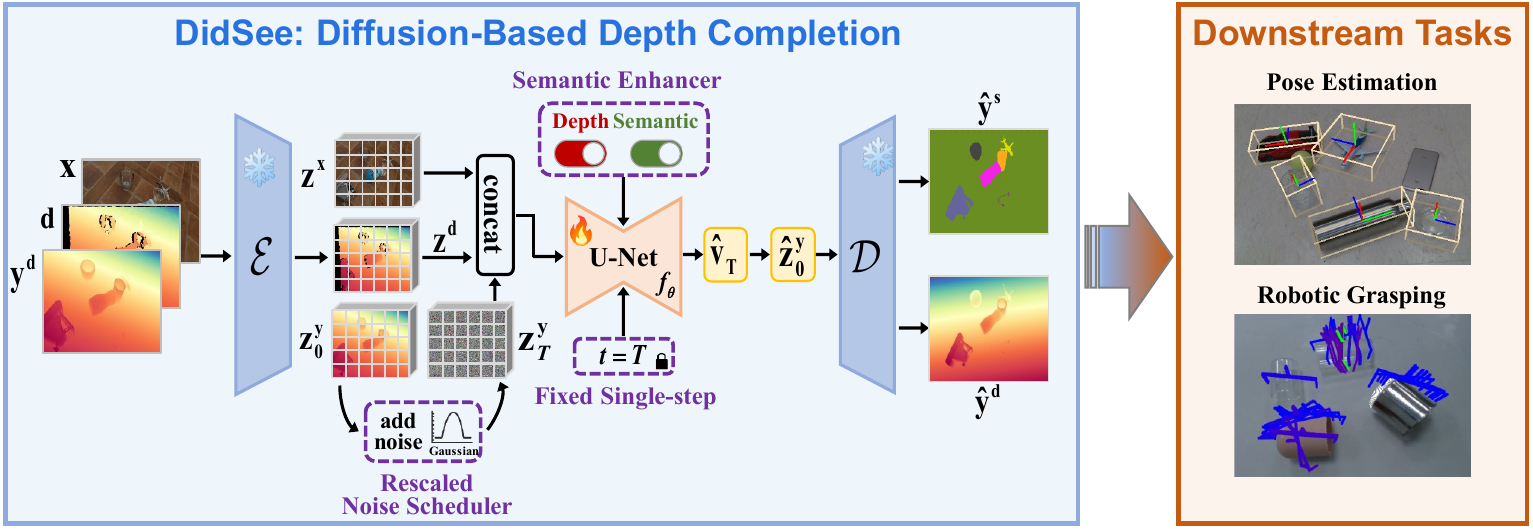}
\vspace{-7pt}
\caption{The overall framework of DidSee. During training, the pre-trained VAE encoder $\mathcal{E}$ encodes the image $\mathbf{x}$, raw depth $\mathbf{d}$, and ground truth depth $\mathbf{y^{d}}$ into latent space, producing $\mathbf{z^x}$, $\mathbf{z^d}$, and $\mathbf{z^y_0}$, respectively. \ding{172} The noisy input $\mathbf{z}_t^\mathbf{y}$ is generated using a rescaled noise scheduler, which enforces a terminal-SNR of zero to eliminate signal leakage bias (\cref{sec:rescaled_schedule}). \ding{173} We adopt a noise-agnostic single-step diffusion formulation with a fixed timestep $t=T$ to mitigate the exposure bias that arises during multi-step sampling (\cref{sec:single_step}). In this formulation, the model's prediction $\hat{\mathbf{v}}_T$ equals the estimated latent $\hat{\mathbf{z}}_0^\mathbf{y}$, which is then decoded by the VAE decoder $\mathcal{D}$ into a depth map. Consequently, we supervise the denoising model $f_{\theta}$ using a task-specific loss in pixel space to enhance performance (\cref{sec:loss_func}). \ding{174} We introduce a novel semantic enhancer that enables the model to jointly perform depth completion and semantic regression, improving object-background distinction and ensuring fine-grained depth prediction (\cref{sec:semantic_enhancer}). \ding{175} The restored depth maps can be applied to downstream tasks, such as pose estimation and robotic grasping on non-Lambertian objects.}
\vspace{-7pt}
\label{fig:overview}
\end{figure*}

%% file: tables/table_noise_scheduler.tex
\begin{table}[!t]
\footnotesize
\centering
\setlength{\tabcolsep}{2pt}
\renewcommand\arraystretch{0.9}
{
\begin{tabular}{c|c|c}
\toprule
Inference Steps  & leading timestep selection & trailing timestep selection\\
\midrule
{1} & {[1]} & {[1000]} \\
{2} & {[501, 1]} &  {[1000, 500]} \\
{5} & {[801, 601, 401, 201, 1]} & {[1000, 800, 600, 400, 200]} \\
\bottomrule   
\end{tabular}}
\vspace{-7pt}
\caption{Different timestep selection strategies with varying numbers of inference steps.}
\vspace{-12pt}
\label{tab:timestep_selection}
\end{table}

%% file: sec/4_method.tex
\section{Methodology}
We address two fundamental biases in standard diffusion models for depth completion, while tackling challenges posed by non-Lambertian surfaces. Our proposed DidSee, illustrated in \cref{fig:overview}, integrates three key components: (1) a rescaled noise scheduler with zero terminal-SNR to mitigate signal leakage bias (\cref{sec:rescaled_schedule}), (2) a noise-agnostic, single-step training formulation to alleviate exposure bias (\cref{sec:single_step}) and incorporate a task-specific loss for improved accuracy (\cref{sec:loss_func}), and (3) a novel semantic enhancer that jointly performs depth completion and semantic segmentation, enabling clearer object-background differentiation and leading to more precise depth predictions (\cref{sec:semantic_enhancer}).
\label{sec:method}
\subsection{Zero Terminal-SNR Noise Scheduler}
\label{sec:rescaled_schedule}
The trailing timestep selection strategy mitigates noise level misalignment during sampling. However, discrepancies between training and inference noise levels persist due to inappropriate noise schedule design.

\para{Signal Leakage Bias.} 
In Stable Diffusion's default configuration ($T=1000$), the noise schedule yields a terminal value of $\bar{\alpha}_T = 0.00466$. At timestep $T$, the corrupted sample $\mathbf{z^y_T}$ follows:
\vspace{-3pt}\begin{equation}\vspace{-3pt}
\label{eqn:corrupted_latent}
	\mathbf{z^y_T} = 0.068265 \cdot \mathbf{z^y_0} + 0.997667 \cdot \boldsymbol{\epsilon}.
\end{equation}
\cref{eqn:corrupted_latent} reveals a subtle yet critical flaw: $\mathbf{z^y_T}$ retains a faint clean signal ($0.068265 \cdot \mathbf{z^y_0}$) at the final timestep. To quantify this, we compute the SNR~\cite{Lin_2024_WACV} as:
\vspace{-3pt}\begin{equation}\vspace{-3pt}
	\label{eqn:snr}
	SNR(t):=\frac{\bar{\alpha}_t}{1-\bar{\alpha}_t}.
\end{equation}
Substituting $\bar{\alpha}_T$ into \cref{eqn:snr} results in $SNR(T) = 0.004682$, indicating that the terminal-SNR is still far from zero. This suggests the presence of signal leakage~\cite{Everaert_2024_WACV} at timestep $T$, creating a mismatch between training and inference. During training, the model expects signal leakage in the corrupted sample $\mathbf{z^y_T}$, whereas at inference, $\mathbf{z^y_T}$ consists solely of Gaussian noise. This signal leakage bias causes errors at the first inference step, which propagate through the subsequent steps, ultimately degrading the model performance. 

\para{Rescaled Noise Scheduler.} 
Signal leakage arises from a nonzero terminal-SNR, \ie $\bar{\alpha_T}\neq0$. To eliminate this, we adopt a rescaled noise scheduler~\cite{Lin_2024_WACV}, which adjusts the noise schedule by keeping $\sqrt{\bar{\alpha_1}}$ unchanged, changing $\sqrt{\bar{\alpha_T}}$ to zero, and linearly rescaling $\sqrt{\bar{\alpha_t}}$ for intermediate $t \in [2, \dots, T-1]$. \cref{fig:rescaled_schedule} compares the original and rescaled noise schedulers on $\bar{\alpha_t}$ and $\text{logSNR}(t)$. This adjustment enforces a zero terminal-SNR, fully aligning the noise levels during inference with the intended training noise at the last timestep $T$ and eliminating signal leakage bias. Though the effect of signal leakage bias is minor in text-to-image generation~\cite{Everaert_2024_WACV}, it proves critical for depth completion, where the rescaled scheduler reduces prediction errors at each inference step, as demonstrated in \cref{fig:exposure_bias}.

\input{figs/fig_rescaled_scheduler}
\input{figs/fig_exposure_bias}

\subsection{Noise-Agnostic Single-Step Training}
\label{sec:single_step}
In addition to mismatches at the terminal timestep, input mismatches also arise at intermediate timesteps $t<T$, potentially accumulating errors during multi-step inference.

\para{Exposure Bias.} The denoising model is trained exclusively with corrupted ground truth samples, $\mathbf{z}_{t}^\mathbf{y}$, rather than with corrupted model-predicted samples, $\hat{\mathbf{z}}_{t}^\mathbf{y}$. This training-inference discrepancy, referred to as exposure bias~\cite{ning_2024_ICLR, Li_2024_ICLR, Li_2024_ICLR_2}, produces additional errors that can progressively accumulate during multi-step inference. To analyze exposure bias in depth completion, we compute the root mean squared error between the predicted latent and the ground truth latent at each step for a fixed number of sampling steps. As shown in \cref{fig:exposure_bias}, since the model inevitably makes errors at each step, these errors propagate through subsequent steps, resulting in steady error accumulation. This error accumulation demonstrates that increasing the number of sampling steps will only exacerbate performance degradation caused by exposure bias in depth completion.

\para{Single-Step Formulation.}
A natural solution to mitigate exposure bias is to perform inference in a single step. Recent studies have reimagined diffusion as a multi-task learning framework~\cite{hang_2023_ICCV}, where each task corresponds to a denoising task at a specific noise level. However, optimizing these tasks jointly can lead to conflicts, as different noise levels may impose competing objectives~\cite{hang_2023_ICCV,go_2023_NIPS,park_2024_ECCV,zheng_2024_ECCV}. To circumvent such conflicts, we train the denoising model in a single-step formulation by fixing $t=T$. Substituting $t=T$ into \cref{eqn:v_target}, the model's prediction target $\mathbf{v}_T$ becomes:
\vspace{-3pt}\begin{equation}\vspace{-3pt}
	\label{eqn:v_target_T}
	\mathbf{v_T} = \sqrt{\bar{\alpha}_T} \boldsymbol{\epsilon} + \sqrt{1-\bar{\alpha}_T} \mathbf{z^y_0}.
\end{equation}
By enforcing a zero terminal-SNR (\cref{sec:rescaled_schedule}), we have $\bar{\alpha}_T = 0$, further simplifying the prediction target to $\mathbf{v_T} = \mathbf{z^y_0}$. As a result, the model's output is no longer dependent on the input noise, and the restored depth map is obtained directly via decoding, i.e., $\hat{\mathbf{y}} = \mathcal{D}(\hat{\mathbf{v}}_T)$. This single-step formulation enables direct optimization using a task-specific loss (\cref{sec:loss_func}) in pixel space, rather than latent space, leading to enhanced depth completion performance.

\subsection{Semantic Enhancer}
\label{sec:semantic_enhancer}
Despite the above modifications, the model still struggles with transparent surfaces due to their lack of distinct visual features, as shown in \cref{fig:semantic_enhancer}. To address this, we introduce a semantic enhancer that leverages semantic segmentation to improve object-background differentiation, enabling more precise depth predictions.

\para{Color Palette.} Since VAE can only map continuous color images, Stable Diffusion~\cite{Stable-Diffusion} cannot directly process discrete semantic segmentation maps due to the inherent differences between regression and classification. To address this, we employ a color palette~\cite{MADM} technique that transforms discrete semantic labels into meaningful RGB images using a predefined palette. These transformed images encode semantic information in a regression-compatible format, allowing seamless integration with diffusion models.

\para{Network Modification.} We utilize task switchers~\cite{Geowizard,Wonder3D,Lotus} to enable the denoiser model $f_\theta$ to jointly perform depth completion and semantic segmentation on the input RGB-D pairs. Each switcher acts as an additional conditioning mechanism, represented as a one-dimensional vector encoded via positional encoding and added to the time embeddings of U-Net~\cite{Wonder3D}, enabling seamless task switching.
\input{figs/fig_enhancer}
To promote task interaction and improve mutual guidance, we further replace the self-attention layers in the U-Net with cross-task self-attention layers~\cite{Geowizard}. This mechanism captures contextual relationships between tasks, enhancing depth completion by integrating semantic knowledge. A detailed description of the modified network is provided in the supplementary material.
\subsection{Loss Functions}
\label{sec:loss_func}
We train DidSee using a combination of depth completion and semantic segmentation losses. The total loss is formulated as follows:
\vspace{-3pt}\begin{equation}\vspace{-3pt}
	\label{eqn:total_loss}
	\mathcal{L}=\left\|\mathbf{\hat{y}^d}-\mathbf{{y}^d}\right\|^1+\gamma\left\|\mathbf{\hat{y}^s}-\mathbf{{y}^s}\right\|^1,
\end{equation}
where the first term represents the depth loss, and the second term corresponds to the semantic segmentation loss. Here, $\mathbf{\hat{y}^d}$ and $\mathbf{y^d}$ denote the predicted and ground truth depth maps, respectively, while $\mathbf{\hat{y}^s}$ and $\mathbf{y^s}$ represent the predicted and ground truth semantic images. $\|\cdot\|^1$ denotes the L1 norm function. The balancing coefficient $\gamma$ is empirically set to 0.1 across all datasets.

%% file: figs/fig_rescaled_scheduler.tex
\begin{figure}[!t]
\centering
  \begin{subfigure}{0.49\columnwidth}
    \includegraphics[width=\linewidth]{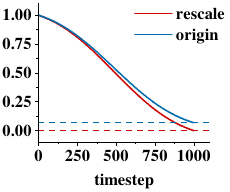}
    \caption{$\sqrt{\bar{\alpha_t}}$}
    \label{fig:sub1}
  \end{subfigure}
  \begin{subfigure}{0.49\columnwidth}
    \includegraphics[width=\linewidth]{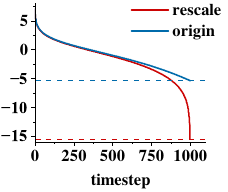}
    \caption{$\text{logSNR}(t)$}
    \label{fig:sub2}
  \end{subfigure}
  \vspace{-7pt}
  \caption{Comparison of original and rescaled noise schedules. (a) Evolution of $\sqrt{\bar{\alpha_t}}$ over timesteps. (b) Evolution of $\text{logSNR}(t)$ over timesteps. The rescaled noise schedule (red) modifies the original schedule (blue) to ensure  $\sqrt{\bar{\alpha_t}}=0$ at the final timestep and attain a terminal-SNR of zero.}
  \vspace{-7pt}
\label{fig:rescaled_schedule}
\end{figure}

%% file: figs/fig_exposure_bias.tex
\begin{figure}[!t]
	\centering
	\includegraphics[width=0.8\columnwidth]{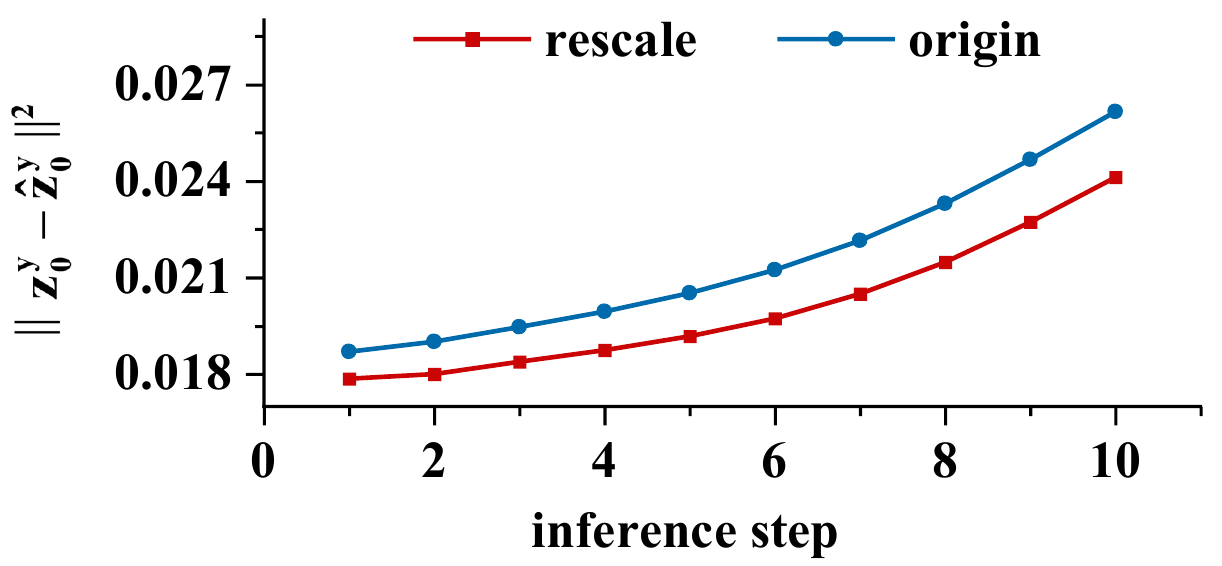}
    \vspace{-7pt}
	\caption{Prediction errors for different numbers of inference steps with original and rescaled noise schedulers. The prediction error is reduced after applying the rescaled noise scheduler. Both curves exhibit a consistent pattern: the error accumulates as the number of inference steps increases.}
    \vspace{-7pt}
	\label{fig:exposure_bias}
\end{figure}

%% file: figs/fig_enhancer.tex
\begin{figure}[!tbp]
	\centering
	\includegraphics[width=1.0\columnwidth]{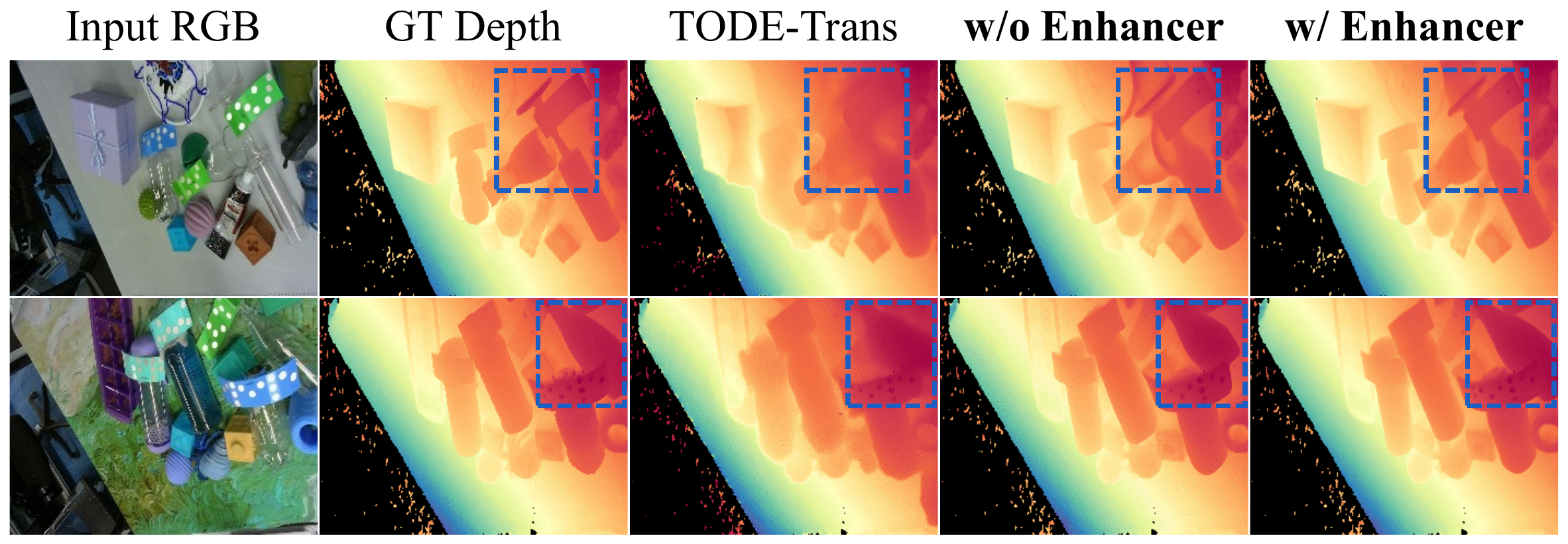}
    \vspace{-7pt}
	\caption{Depth maps without and with semantic enhancer. Models typically generate accurate depth in opaque regions but are prone to errors on transparent surfaces.}
    \vspace{-7pt}
	\label{fig:semantic_enhancer}
\end{figure}

%% file: sec/5_exp.tex
\section{Experiments}
We conduct extensive experiments to evaluate DidSee across multiple benchmarks and compare its performance against state-of-the-art methods. Additionally, we perform ablation studies to analyze the contributions of different components in our approach. To further assess the quality of the restored depth maps, we apply them to two downstream tasks: category-level pose estimation and robotic grasping.
\subsection{Datasets and Metrics}
\para{TransCG} dataset~\cite{DFNet} consists of 57K RGB-D images captured by two different cameras, featuring 51 transparent objects across 130 real-world scenes.

\para{ClearPose} dataset~\cite{ClearPose} includes over 350K labeled real-world RGB-D frames, covering 63 household objects. It presents challenging test scenarios such as heavy occlusions, non-planar orientations, and a translucent cover.

\para{DREDS \& STD} dataset~\cite{SwinDR} comprises both synthetic and real annotated data for objects with diverse material properties (specular, transparent, and diffuse). DREDS is a large-scale synthetic dataset containing 130K photorealistic RGB-D images. STD is a real-world dataset that contains 27K RGB-D frames from 30 cluttered scenes with 50 objects. Both datasets are divided into \textit{CatKnown} and \textit{CatNovel}, based on the novelty of object categories.

\para{Metrics.} 
Following prior research~\cite{ClearGrasp,DFNet,FDCT,TODE-Trans,SwinDR}, we assess the depth completion performance on non-Lambertian objects using standard metrics: root mean square error (RMSE), relative depth error (REL), mean absolute error (MAE), along with accuracy within threshold $\delta_{\tau}$ (with $\tau$ set to 1.05, 1.10, and 1.25).

\input{tables/table1}
\subsection{Implementation Details} 
We implement DidSee using PyTorch and initialize it with pre-trained weights from Stable Diffusion v2~\cite{Stable-Diffusion}. During training, we fix the timestep $t=1000$. We train DidSee for 40 epochs with a batch size of 16, using the AdamW optimizer with a learning rate of $5\cdot 10^{-5}$. For a fair comparison, the images are randomly cropped to $320 \times 240$ for the TransCG and DREDS datasets, and $640 \times 480$ for the ClearPose dataset. All experiments are conducted on 8 NVIDIA A100 GPUs.

\subsection{Quantitative Results}
\para{TransCG.} We first evaluate DidSee on the TransCG dataset (\cref{tab:transcg}). DidSee surpasses all competing methods across every evaluation metric. In particular, compared to the second-best method, TODE-Trans~\cite{TODE-Trans}, DidSee achieves notable improvements of 15.4\%, 26.3\%, and 25.0\% for RMSE, REL, and MAE, respectively.

\para{ClearPose.} 
ClearPose presents a greater challenge due to its diverse testing scenarios. As shown in \cref{tab:ClearPose}, DidSee surpasses the baselines across nearly all test scenes, highlighting its strong \textbf{scene generalization}. While all methods generally perform unfavorably in the Translucent Cover scene, DidSee outperforms competitors by a significant margin (\eg, 40.0\% on REL over DFNet~\cite{DFNet} and 53.5\% on $\delta_{1.05}$ over LIDF~\cite{LIDF}).

\input{tables/table2}
\para{DREDS \& STD.} 
\cref{tab:DREDS} presents the quantitative results on DREDS and STD datasets. DidSee consistently achieves the lowest error rates and the highest threshold-based accuracy metrics across both synthetic and real-world subsets. The results on DREDS-CatNovel highlight DidSee's strong \textbf{cross-category generalization}, while its performance on STD-CatKnown demonstrates robust \textbf{sim-to-real generalization}. Furthermore, its success on STD-CatNovel indicates that DidSee effectively addresses both challenges simultaneously, reinforcing its adaptability to real-world scenarios. We also provide a full-depth comparison with D3RoMa~\cite{D3roma}, a diffusion-based stereo depth estimation framework that also supports RGB-D inputs.
As shown in \cref{tab:comp_d3roma}, DidSee consistently outperforms D3RoMa~\cite{D3roma} and TODE-Trans~\cite{TODE-Trans}, particularly on the real-world dataset STD-CatNovel.

\input{tables/table3}
\input{figs/fig_dreds}
\input{figs/fig_wild}
\subsection{Qualitative Results}
To further validate the effectiveness of DidSee, we compare its point cloud reconstructions with those produced by competing methods. As illustrated in \cref{fig:qualitative_dreds}, DidSee exhibits superior sim-to-real transferability by generating fewer artifacts, and more precise scene reconstructions. These improvements are especially notable in challenging regions characterized by strong specular reflections or transparent surfaces. We also provide qualitative comparisons on real-world, in-the-wild scenes. In \cref{fig:qualitative_wild}, DidSee delivers sharper depth boundaries and more complete object shapes. Benefiting from the extensive visual priors within the pre-trained diffusion model, DidSee effectively generalizes to diverse real-world conditions, confirming its robustness in the wild.
\input{tables/supp_table1}
\input{tables/ablate_component}
\subsection{Ablation Studies}
We perform a series of ablation experiments to examine how each proposed component affects the performance of DidSee, as summarized in \cref{tab:ablate_cg}. We start with a baseline model that applies a standard diffusion framework with trailing timestep selection strategy. First, we adopt the zero terminal-SNR noise scheduler (\cref{sec:rescaled_schedule}), which reduces REL by over 20\% and reveals the detrimental impact of signal leakage bias on depth completion. Next, we integrate noise-agnostic single-step training (\cref{sec:single_step}), yielding further gains by reducing REL from 0.020 to 0.017. Combined with the findings in \cref{fig:exposure_bias}, these results underscore the adverse effect of exposure bias in multi-step formulations. Finally, we integrate the semantic enhancer (\cref{sec:semantic_enhancer}), which refines depth predictions by enhancing object-background distinction. With all components combined, the complete DidSee model achieves the best performance across all evaluation metrics.

\subsection{Downstream Robotic Task Evaluation}
We evaluate the restored depth maps on two robotic downstream tasks: category-level pose estimation and robotic manipulation. Each task involves a diverse set of objects with varying material properties, including diffuse, transparent, and specular surfaces. For additional experimental details, please refer to the supplementary material. The scene reconstruction and robotic grasping demos can be found in the supplementary video.

\input{tables/table4}
\input{tables/table5}
\para{Category-level Pose Estimation.}
We conduct evaluations following the protocol in~\cite{SwinDR}. The pose estimation model uses restored depth maps from various methods for pose fitting. We report two metrics for category-level pose estimation:
\begin{itemize}
	\item $\text{IoU}_e$: The mean average precision for the intersection over union of 3D bounding boxes with a threshold of $e$.
	\item $n ^\circ m \ \text{cm}$: The mean average precision for objects with translation errors less than $m$ cm and rotation errors less than $n^\circ$.
\end{itemize}
As shown in \cref{tab:pose_estimation}, DidSee excels in category-level pose estimation, achieving superior performance across nearly all metrics. These results demonstrate the high quality of depth maps generated by DidSee.

\para{Robotic Grasping.}
To assess depth restoration in robotic manipulation, we integrate various depth restoration methods into the GraspNet~\cite{GraspNet} pipeline. We quantify real-world robotic grasping performance using the mean percentage of successfully removed objects across all items. Quantitative results in \cref{tab:grasping} show that the improved depth quality from DidSee leads to the highest success rates in grasp execution and the most reliable manipulation outcomes, especially for specular and transparent objects. These results emphasize the broader applicability of DidSee in material-agnostic robotic applications.

%% file: tables/table1.tex
\begin{table}[!t]
	\footnotesize
	\centering
	\setlength{\tabcolsep}{3pt}
	\renewcommand\arraystretch{0.87}
	\begin{tabular}{l|cccccc}
		\toprule
		
		Methods
		& RMSE$\downarrow$ 
		& REL$\downarrow$ 
		& MAE$\downarrow$ 
		& $\delta_{1.05}$$\uparrow$ 
		& $\delta_{1.10}$$\uparrow$ 
		& $\delta_{1.25}$$\uparrow$ \\
		
		\midrule
		
		{ClearGrasp~\cite{ClearGrasp}} & {0.054} & {0.083} & {0.037} & {50.48} & {68.68} & {95.28} \\
		{LIDF~\cite{LIDF}} & {0.019} & {0.034} & {0.015} & {78.22} & {94.26} & {99.80} \\
		{TranspareNet~\cite{TranspareNet}} & {0.026} & {0.023} & {0.013} & {88.45} & {96.25} & {99.42} \\
		{DFNet~\cite{DFNet}} & {0.018} & {0.027} & {0.012} & {83.76} & {95.67} & {99.71} \\
		{TCRNet~\cite{TCRNet}} & {0.017} & {0.020} & {0.010} & {88.96} & {96.94} & \second{99.87} \\
		{FDCT~\cite{FDCT}} & {0.015} & {0.022} & {0.010} & {88.18} & {97.15} & {99.81} \\
		{TODE-Trans~\cite{TODE-Trans}} & \second{0.013} & \second{0.019} & \second{0.008} & \second{90.43} & \second{97.39} & {99.81} \\
		
		\midrule
		{DidSee (Ours)} & {\first {0.011}} & {\first {0.014}} & {\first {0.006}} & {\first {94.19}} & {\first {98.33}} & {\first {99.91}} \\
		\bottomrule   
	\end{tabular}
    \vspace{-7pt}
	\caption{Quantitative comparison on TransCG dataset. The \colorbox{red!15}{\textbf{best}} and \colorbox{blue!15}{second-best} results are highlighted. $\uparrow$ means higher is better and $\downarrow$ means lower is better.}
    \vspace{-3pt}
\label{tab:transcg}
\end{table}
%

%% file: tables/table2.tex


{\setlength{\tabcolsep}{3pt}
\renewcommand\arraystretch{0.87}
\begin{table}[!t]
\footnotesize
\centering
\begin{tabular}{l|cccccc}
\toprule
{Methods} 
& RMSE$\downarrow$ 
& REL$\downarrow$ 
& MAE$\downarrow$ 
& $\delta_{1.05}$$\uparrow$ 
& $\delta_{1.10}$$\uparrow$ 
& $\delta_{1.25}$$\uparrow$ 
\\

\midrule

{} & \multicolumn{6}{c}{New BackGround} \\
\midrule
{LIDF~\cite{LIDF}} & {0.07} & {0.05} & {0.04} & {67.00} & {87.03} & {97.50} \\
{DFNet~\cite{DFNet}} & \second{0.03} & {0.03} & \second{0.02} & {86.50} & {97.02} & {99.74} \\
{FDCT$^*$~\cite{FDCT}} & \second{0.03} & \second{0.02} & \second{0.02} & {90.99} & {97.56} & \second{99.61} \\
{TODE-Trans$^*$~\cite{TODE-Trans}} & \second{0.03} & \second{0.02} & \second{0.02} & \second{91.62} & \second{97.68} & {99.57} \\
\cmidrule{1-7}
{DidSee (Ours)} & \first{0.02} & \first{0.01} & \first{0.01} & \first{96.12} & \first{98.56} & \first{99.69} \\

\midrule
{} & \multicolumn{6}{c}{Heavy Occlusion} \\
\midrule

{LIDF~\cite{LIDF}} & {0.11} & {0.09} & {0.08} & {41.43} & {66.52} & {91.96} \\
{DFNet~\cite{DFNet}} & {0.06} & \second{0.04} & {0.04} & {72.03} & {90.61} & {98.73} \\
{FDCT$^*$~\cite{FDCT}} & \second{0.05} & \first{0.03} & \second{0.03} & \second{81.09} & \second{93.74} & \second{99.08} \\
{TODE-Trans$^*$~\cite{TODE-Trans}} & {0.05} & \second{0.04} & \second{0.03} & {79.35} & {92.19} & {98.67} \\
\cmidrule{1-7}
{DidSee (Ours)} & \first{0.04} & \first{0.03} & \first{0.02} & \first{85.59} & \first{94.91} & \first{99.12} \\

\midrule
{} & \multicolumn{6}{c}{Translucent Cover} \\
\midrule

{LIDF~\cite{LIDF}} & \second{0.16} & {0.16} & \second{0.13} & {22.85} & \second{41.17} & \second{73.11} \\
{DFNet~\cite{DFNet}} & \second{0.16} & \second{0.15} & {0.14} & \second{23.44} & {39.75} & {67.56} \\
{FDCT$^*$~\cite{FDCT}} & {0.17} & {0.17} & {0.14} & {20.77} & {34.92} & {62.69} \\
{TODE-Trans$^*$~\cite{TODE-Trans}} & {0.17} & {0.17} & {0.15} & {18.51} & {32.39} & {62.38} \\
\cmidrule{1-7}
{DidSee (Ours)} & \first{0.10} & \first{0.09} & \first{0.08} & \first{35.98} & \first{61.94} & \first{91.23} \\

\midrule
{} & \multicolumn{6}{c}{Opaque Distractor} \\
\midrule

{LIDF~\cite{LIDF}} & {0.14} & {0.13} & {0.10} & {34.41} & {55.59} & {83.23} \\
{DFNet~\cite{DFNet}} & {0.08} & \second{0.06} & {0.06} & {52.43} & {75.52} & {97.53} \\
{FDCT$^*$~\cite{FDCT}} & \second{0.07} & \first{0.05} & \first{0.04} & \second{66.69} & \second{85.05} & \first{98.00} \\
{TODE-Trans$^*$~\cite{TODE-Trans}} & {0.08} & {0.06} & \second{0.05} & {61.37} & {79.85} & {95.98} \\
\cmidrule{1-7}
{DidSee (Ours)} & \first{0.06} & \first{0.05} & \first{0.04} & \first{70.79} & \first{87.94} & \second{97.84} \\

\midrule
{} & \multicolumn{6}{c}{Filled Liquid} \\
\midrule

{LIDF~\cite{LIDF}} & {0.14} & \second{0.13} & \second{0.11} & {32.84} & {53.44} & {84.84} \\
{DFNet~\cite{DFNet}} & \first{0.04} & \first{0.04} & \first{0.03} & {77.65} & \first{93.81} &  \first{99.50} \\
{FDCT$^*$~\cite{FDCT}} & \second{0.05} & \first{0.04} & \first{0.03} & \second{79.52} & \second{93.70} & \second{99.07} \\
{TODE-Trans$^*$~\cite{TODE-Trans}} & \second{0.05} & \first{0.04} & \first{0.03} & 74.54{} & {91.27} & {98.91} \\
\cmidrule{1-7}
{DidSee (Ours)} & \first{0.04} & \first{0.04} & \first{0.03} & \first{80.82} & {93.44} & {98.90} \\

\midrule
{} & \multicolumn{6}{c}{Non Planar} \\
\midrule

{LIDF~\cite{LIDF}} & {0.18} & {0.16} & {0.15} & {20.34} & {38.57} & {74.02} \\
{DFNet~\cite{DFNet}} & {0.09} & {0.07} & {0.07} & {55.31} & {76.47} & {94.88} \\
{FDCT$^*$~\cite{FDCT}} & \second{0.07} & \second{0.06} & \second{0.05} & {64.44} & \second{83.91} & \second{96.75} \\
{TODE-Trans$^*$~\cite{TODE-Trans}} & {0.08} & {0.06} & {0.05} & \second{64.7} & {83.69} & {96.12} \\
\cmidrule{1-7}
{DidSee (Ours)} & \first{0.06} & \first{0.04} & \first{0.04} & \first{76.18} & \first{89.62} & \first{98.06} \\

\bottomrule
\end{tabular}
\vspace{-7pt}
\caption{Quantitative comparison on ClearPose dataset. $^*$ indicates results reproduced by us.}
\vspace{-9pt}
\label{tab:ClearPose}
\end{table}}

%% file: tables/table3.tex


\begin{table*}[t]
\footnotesize
\centering
\renewcommand\arraystretch{0.8}
{
\begin{tabular}{l|cccccc}
\toprule

{Methods} 
& RMSE$\downarrow$ 
& REL$\downarrow$ 
& MAE$\downarrow$ 
& $\delta_{1.05}$$\uparrow$ 
& $\delta_{1.10}$$\uparrow$ 
& $\delta_{1.25}$$\uparrow$ 
\\

\midrule
\midrule
{} & \multicolumn{6}{c}{DREDS-CatKnown (Sim)} \\
\midrule
{LIDF~\cite{LIDF}} & {0.016 / 0.015} & {0.018 / 0.017} & {0.011 / 0.011} & {93.60 / 94.45} & {98.71 / 98.79} & {99.92 / 99.90} \\
{DFNet$^*$~\cite{DFNet}} & {0.020 / 0.022} & {0.021 / 0.025} & {0.013 / 0.016} & {88.77 / 85.62} & {97.48 / 96.83} & {99.87 / 99.83} \\
{FDCT$^*$~\cite{FDCT}} & {0.012 / 0.013} & {0.011 / 0.013} & {0.007 / 0.008} & {96.46 / 95.55} & {99.28 / 99.13} & {99.96 / 99.95} \\
{SWinDRNet~\cite{SwinDR}} & {0.010 / 0.010} & {0.008 / 0.009} & {0.005 / 0.006} & {98.04 / 97.76} & {99.62 / 99.57} & \second{99.98 / 99.97} \\
{TODE-Trans$^*$~\cite{TODE-Trans}} & \second{0.007 / 0.008} & \second{0.006 / 0.007} & \second{0.004 / 0.004} & \second{98.91 / 98.65} & \second{99.77 / 99.70} & \second{99.98 / 99.97} \\

\midrule

{DidSee (Ours)} & \first {0.005 / 0.005} & \first {0.004 / 0.004} & \first {0.002 / 0.003} & \first {99.50 / 99.39} & \first {99.88 / 99.85} & \first {99.99 / 99.98}  \\

\midrule
{} & \multicolumn{6}{c}{DREDS-CatNovel (Sim)} \\
\midrule
{LIDF~\cite{LIDF}} & {0.082 / 0.082} & {0.183 / 0.184} & {0.069 / 0.069} & {23.70 / 23.69} & {42.77 / 42.88} & {75.44 / 75.54} \\
{DFNet$^*$~\cite{DFNet}} & {0.031 / 0.035} & {0.054 / 0.067} & {0.021 / 0.023} & {69.53 / 60.65} & {85.73 / 80.51} & {97.13 / 96.29} \\
{FDCT$^*$~\cite{FDCT}} & {0.023 / 0.027} & {0.036 / 0.047} & {0.014 / 0.018} & {80.67 / 73.09} & {91.40 / 87.83} & {98.28 / 97.66} \\
{SWinDRNet~\cite{SwinDR}} & {0.022 / 0.025} & {0.034 / 0.044} & {0.013 / 0.017} & {81.90 / 75.27} & {92.18 / 89.15} & {98.39 / 97.81} \\
{TODE-Trans$^*$~\cite{TODE-Trans}} & \second{0.016 / 0.019} & \second{0.023 / 0.031} & \second{0.009 / 0.012} & \second{87.51 / 82.33} & \second{95.18 / 93.17} & \second{99.05 / 98.63} \\

\midrule
{DidSee (Ours)} & \first {0.015 / 0.017} & \first {0.021 / 0.028} & \first {0.008 / 0.011} & \first {88.83 / 83.86} & \first {95.67 / 93.82} & \first {99.24 / 98.95}  \\

\midrule
{} & \multicolumn{6}{c}{STD-CatKnown (Real)} \\
\midrule
{LIDF~\cite{LIDF}} & {0.019 / 0.022} & {0.019 / 0.023} & {0.013 / 0.015} & {93.08 / 90.32} & {98.39 / 97.38} & {99.83 / 99.62} \\
{DFNet$^*$~\cite{DFNet}} & {0.029 / 0.041} & {0.028 / 0.048} & {0.018 / 0.031} & {82.73 / 62.07} & {95.80 / 91.20} & {99.68 / 99.37} \\
{FDCT$^*$~\cite{FDCT}} & {0.017 / 0.022} & {0.016 / 0.023} & {0.010 / 0.015} & {94.86 / 90.40} & {98.89 / 98.29} & {99.91 / 99.81} \\
{SWinDRNet~\cite{SwinDR}} & {0.015 / 0.018} & {0.013 / 0.016} & {0.008 / 0.011} & {96.66 / 94.97} & {99.03 / 98.79} & {99.92 / 99.85} \\
{TODE-Trans$^*$~\cite{TODE-Trans}} & \second{0.013 / 0.016} & \second{0.011 / 0.014} & \second{0.007 / 0.010} & \second{97.81 / 96.23} & \second{99.30 / 98.74} & \first {99.95 / 99.90} \\

\midrule
{DidSee (Ours)} & \first {0.012 / 0.014} & \first {0.009 / 0.011} & \first {0.006 / 0.007} & \first {98.38 / 97.74} & \first {99.36 / 99.08} & \first {99.95 / 99.90}  \\

\midrule
{} & \multicolumn{6}{c}{STD-CatNovel (Real)} \\
\midrule
{LIDF~\cite{LIDF}} & {0.041} & {0.060} & {0.031} & {53.69} & {79.80} & {99.63} \\
{DFNet$^*$~\cite{DFNet}} & {0.034} & {0.049} & {0.025} & {64.81} & {87.22} & {99.56} \\
{FDCT$^*$~\cite{FDCT}} & {0.028} & {0.037} & {0.019} & {78.91} & {92.02} & {99.60} \\
{SWinDRNet~\cite{SwinDR}} & {0.025} & {0.033} & {0.017} & {81.55} & \second{93.10} & \second{99.84} \\
{TODE-Trans$^*$~\cite{TODE-Trans}} & \second{0.024} & \second{0.030} & \second{0.016} & \second{84.59} & {92.37} & \first {99.94} \\

\midrule
{DidSee (Ours)} & \first {0.022} & \first {0.026} & \first {0.014} & \first {85.00} & \first {92.72} & \first {99.94}  \\

\bottomrule
\end{tabular}
}
\vspace{-8pt}
\caption{Quantitative comparison on DREDS and STD (all objects / non-Lambertian objects). All methods are trained on the DREDS-CatKnown training split. Notably, every object in STD-CatNovel is non-Lambertian.}
\vspace{-7pt}
\label{tab:DREDS}
\end{table*}

%% file: figs/fig_dreds.tex
\begin{figure*}[!htbp]
	\centering
	\includegraphics[width=0.9\linewidth]{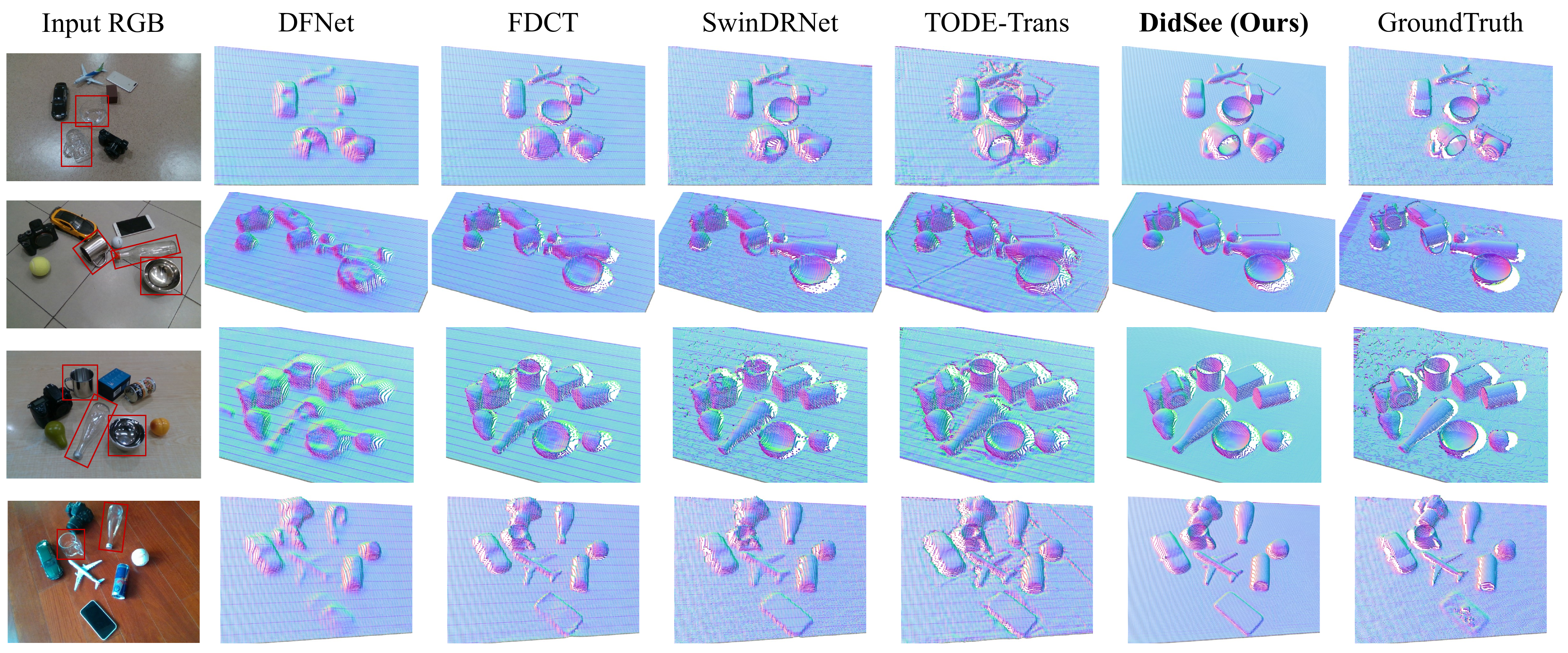}
    \vspace{-10pt}
	\caption{Qualitative comparison on STD-CatKnown dataset. Red bounding boxes delineate challenging non-Lambertian objects.
    }
    \vspace{-7pt}
	\label{fig:qualitative_dreds}
\end{figure*}

%% file: figs/fig_wild.tex
\begin{figure*}[!htbp]
	\centering
	\includegraphics[width=0.9\linewidth]{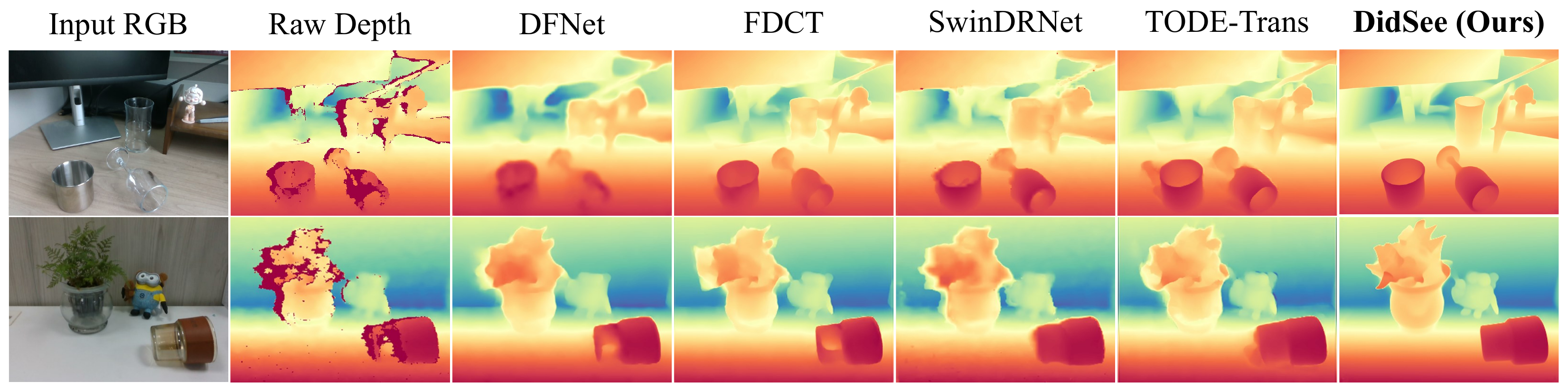}
    \vspace{-12pt}
	\caption{Qualitative comparison on in-the-wild scenes.}
    \vspace{-18pt}
	\label{fig:qualitative_wild}
\end{figure*}

%% file: tables/supp_table1.tex

\begin{table}[!t]
\centering
\setlength{\tabcolsep}{6pt}
\renewcommand\arraystretch{0.88}
\resizebox{1\linewidth}{!}
{
\begin{tabular}{l|cccccc}
\toprule

{Methods} 
& RMSE$\downarrow$ 
& REL$\downarrow$ 
& MAE$\downarrow$ 
& $\delta_{1.05}$$\uparrow$ 
& $\delta_{1.10}$$\uparrow$ 
& $\delta_{1.25}$$\uparrow$ 
\\

\midrule

{} & \multicolumn{6}{c}{STD-CatKnown (Real)} \\
\midrule
{TODE-Trans*~\cite{TODE-Trans}} & {0.0122} & {0.0050} & {0.0037} & {98.20} & {99.37} & \second{99.93} \\
{D3RoMa~\cite{D3roma}} & \second{0.0101} & \second{0.0042} & \second{0.0030} & \second{99.03} & \second{99.57} & \second{99.93} \\
\midrule
{Didsee (Ours)} & \first{0.0091} & \first{0.0039} & \first{0.0028} & \first{99.18} & \first{99.66} & \first{99.96}  \\

\midrule
{} & \multicolumn{6}{c}{STD-CatNovel (Real)} \\
\midrule
{TODE-Trans*~\cite{TODE-Trans}} & \second{0.0147} & \second{0.0089} & \second{0.0053} & \second{95.24} & \second{97.95} & \second{99.91} \\
{D3RoMa~\cite{D3roma}} & {0.0397} & {0.0158} & {0.0092} & {92.78} & {97.13} & {99.61} \\
\midrule
{Didsee (Ours)} & \first{0.0113} & \first{0.0060} & \first{0.0034} & \first{97.23} & \first{98.68} & \first{99.92}  \\
\bottomrule
\end{tabular}
}
\vspace{-7pt}
\caption{Quantitative comparison on STD dataset for complete depth map evaluation.} 
\vspace{-7pt}
\label{tab:comp_d3roma}
\end{table}

%% file: tables/ablate_component.tex
\begin{table}[!t]
	\centering
        \setlength{\tabcolsep}{1.6pt}
    \renewcommand\arraystretch{0.87}
        \resizebox{ \linewidth}{!}
	{
		\begin{tabular}{l|c|cccccc}
			\toprule
			Methods
			& Steps 
			& RMSE$\downarrow$ 
			& REL$\downarrow$ 
			& MAE$\downarrow$ 
			& $\delta_{1.05}$$\uparrow$ 
			& $\delta_{1.10}$$\uparrow$ 
			& $\delta_{1.25}$$\uparrow$ \\
			
			\midrule
			{Baseline} & {10} & {0.017} & {0.026} & {0.012} & {88.65} & {96.24} & {99.69} \\
			{$+$ Zero Terminal-SNR Noise Scheduler} & {10} & {0.014} & {0.020} & {0.009} & {91.60} & {97.05} & {99.76} \\
			{$+$ Noise-Agnostic Single-step Training}& {1} & \second{0.013} & \second{0.017} & \second{0.008} & \second{91.96} & \second{97.10} & \second{99.79} \\
			\midrule
			{$+$ Semantic Enhancer (DidSee)}& {1} & {\first {0.011}} & {\first {0.014}} & {\first {0.006}} & {\first {94.19}} & {\first {98.33}} & {\first {99.91}} \\
			\bottomrule   
		\end{tabular}}
        \vspace{-7pt}
	\caption{Ablation studies on the incremental design of DidSee.}
        \vspace{-7pt}
	\label{tab:ablate_cg}
\end{table}

%% file: tables/table4.tex


\begin{table}[!t]
	\footnotesize
	\centering
	\setlength{\tabcolsep}{2pt}
\renewcommand\arraystretch{0.87}
	\resizebox{1\linewidth}{!}
	{
		
		\begin{tabular}{{l|ccccccc}}
			\toprule
			{Methods} & $\text{IoU}_{25}$ $\uparrow$ & $\text{IoU}_{50}$$\uparrow$ & $\text{IoU}_{75}$ $\uparrow$ & $5^\circ 2$cm$\uparrow$ & $5^\circ 5$cm$\uparrow$ & $10^\circ 2$cm$\uparrow$ & $10^\circ 5$cm$\uparrow$ \\
			
			\midrule
			{} & \multicolumn{7}{c}{DREDS-CatKnown (Sim)} \\
			\midrule
			
			{Raw} & {94.3} & {78.8} & {36.7} & {34.6} & {37.8} & {55.9} & {62.9} \\
			{LIDF~\cite{LIDF}} & {94.6} & {80.7} & {36.6} & {33.9} & {36.4} & {58.2} & {64.7} \\
			{DFNet$^*$~\cite{DFNet}} & {94.5} & {78.9} & {35.2} & {32.4} & {35.3} & {55.1} & {61.1} \\
			{FDCT$^*$~\cite{FDCT}} & {95.1} & {84.2} & {47.2} & {41.9} & {42.8} & {66.7} & {69.3}  \\
			{SWinDRNet~\cite{SwinDR}} & \first{{95.3}} & {85.0} & {49.9} & {49.3} & {50.3} & {70.1} & {72.8} \\
			{TODE-Trans$^*$~\cite{TODE-Trans}} & \first{{95.3}} & \second{85.2} & \second{50.1} & \second{49.7} & \second{50.6} & \second{70.6} & \second{73.1} \\
            \midrule
			{DidSee (Ours)} & \first{{95.3}} & \first{{85.3}} & \first{{50.3}} & \first{{52.2}} & \first{{53.1}} & \first{{71.3}} & \first{{73.8}} \\
			
			\midrule
			{} & \multicolumn{7}{c}{STD-CatKnown (Real)} \\
			\midrule
			
			{Raw} & {91.5} & {81.3} & {39.3} & {38.2} & {42.9} & {58.3} & {71.2} \\
			{LIDF~\cite{LIDF}} & {91.3} & {83.2} & {42.9} & {34.8} & {37.4} & {65.2} & {71.0}  \\
			{DFNet$^*$~\cite{DFNet}} & {91.4} & {79.9} & {35.1} & {32.2} & {36.6} & {52.2} & {63.6} \\
			{FDCT$^*$~\cite{FDCT}} & {91.6} & {85.7} & {53.4} & {49.6} & {50.5} & {75.2} & {77.5} \\
			{SWinDRNet~\cite{SwinDR}} & \second {91.5} & \first {{85.7}} & \first {{55.7}} & {53.3} & {54.1} & {77.6} & {79.7} \\
			{TODE-Trans$^*$~\cite{TODE-Trans}} & \second{91.5} & \first {{85.7}} & {55.0} & \second {54.6} & \second{55.3} & \second{79.3} & \second{81.1} \\
            \midrule
			{DidSee (Ours)} & \first {{91.6}} & \first {{85.7}} & \second {{55.6}} & \first {{56.6}} & \first {{57.7}} & \first {{80.2}} & \first {{82.7}} \\
			\bottomrule   
		\end{tabular}
	}
        \vspace{-7pt}
	\caption{Quantitative comparison for category-level pose estimation. \textit{Raw} refers to using raw depth for pose fitting.}
    \vspace{-7pt}
	\label{tab:pose_estimation}
\end{table}

%% file: tables/table5.tex
\begin{table}[!t]
\centering
\setlength{\tabcolsep}{8pt}
\renewcommand\arraystretch{0.88}
\resizebox{ \linewidth}{!}
{
\begin{tabular}{l|cccc}
\toprule
{Methods} 
&{Transparent}
&{\ Specular\ }
&{\ \  Diffuse\ \ }
&{\ \  Overall\ \  }
\\
\midrule
SWinDRNet~\cite{SwinDR} & 82.4\% & 84.1\% & 91.7\% & 85.5\% \\
TODE-Trans$^*$~\cite{TODE-Trans} & \second{85.1\%} & \second{88.6\%} & \second{95.8\%} & \second{89.2\%} \\
\midrule
DidSee (Ours) & \first{89.2\%} & \first{93.2\%} & \first{97.9\%} & \first{92.8\%}\\
\bottomrule
\end{tabular}
}
\vspace{-7pt}
\caption{Quantitative comparison of robotic grasping performance on objects with different materials.}
\vspace{-7pt}
\label{tab:grasping}
\end{table}

%% file: sec/6_conclusion.tex
\section{Conclusion}
In this work, we propose DidSee, a diffusion-based method for depth restoration of non-Lambertian objects. DidSee effectively mitigates two critical biases inherent in diffusion models for depth completion, substantially reducing prediction errors. Furthermore, the implementation of a semantic enhancer significantly enhances DidSee's ability to differentiate non-Lambertian surfaces from backgrounds, ensuring more precise depth maps. Comprehensive experiments show that DidSee achieves SoTA performance on multiple datasets~\cite{DFNet,ClearPose,SwinDR} and enhances the efficacy of downstream robotic tasks across various object materials.

%% file: sec/7_appendix.tex
\clearpage
\setcounter{page}{1}
\maketitlesupplementary

\appendix
\noindent In this supplementary material, we provide the implementation and experimental details in \cref{supp:implementation} and \cref{supp:experiment}, respectively. We also present additional qualitative results in \cref{supp:qualitative}. Finally, we discuss the limitations of our approach in \cref{supp:limitation}.
\section{Implementation Details}
\label{supp:implementation}
\subsection{Data Preprocessing}
Following previous works~\cite{DFNet,D3roma,FDCT}, we filter out outliers from depth maps. For TransCG~\cite{DFNet} and ClearPose~\cite{ClearPose}, valid depth values fall within the range $[0.3,1.5]$. In DREDS, the depth range is $[0.0,3.5]$, while for STD, it is $[0.2,2]$. Missing depth values are interpolated and subsequently normalized to $[-1,1]$. To match the input format of the VAE encoder, the normalized depth map is replicated across three color channels. Additionally, we apply random flipping and cropping for data augmentation.
\subsection{Network Architectures}
\para{Task Switcher.} We incorporate task switchers~\cite{Wonder3D}, denoted as $\mathbf{m^d}$ and $\mathbf{m^s}$, to enable the denoiser model $f_\theta$ to simultaneously perform the depth completion and semantic regression. Each task switcher is a one-dimensional vector encoded via a positional encoder and added to the time embeddings of the diffusion model. Formally, we define the outputs as:
\vspace{-3pt}\begin{equation}\vspace{-3pt} \hat{\mathbf{y}}^\mathbf{d} = \mathcal{D}\left(f_{\theta}([\mathbf{z}^\mathbf{x},\mathbf{z}^\mathbf{d}],\mathbf{m^d})\right), \ \hat{\mathbf{y}}^\mathbf{s} = \mathcal{D}\left(f_{\theta}([\mathbf{z}^\mathbf{x},\mathbf{z}^\mathbf{d}],\mathbf{m^s})\right), \end{equation}
where $\hat{\mathbf{y}}^\mathbf{d}$ and $\hat{\mathbf{y}}^\mathbf{s}$ correspond to the predicted depth maps and semantic images, respectively. $\mathbf{z}^\mathbf{x}$ and $\mathbf{z}^\mathbf{d}$ denote the input RGB and raw depth latent, while $\mathcal{D}$ represents the VAE decoder.

\para{Cross-Task Attention.} 
To enhance mutual task guidance, we replace the standard self-attention layer in U-Net with a cross-task attention mechanism~\cite{Geowizard}. The queries, keys, and values are computed as:
\vspace{-3pt}
\begin{equation}\vspace{-3pt}
\begin{aligned}
& \mathbf{q}^{\mathbf{d}}=\mathbf{Q} \cdot \hat{\mathbf{z}}^{\mathbf{d}}, \  
\mathbf{k}^{\mathbf{d}}=\mathbf{K} \cdot\left(\hat{\mathbf{z}}^{\mathbf{d}} \oplus \hat{\mathbf{z}}^{\mathbf{s}}\right), \  
\mathbf{v}^{\mathbf{d}}=\mathbf{V} \cdot\left(\hat{\mathbf{z}}^{\mathbf{d}} \oplus \hat{\mathbf{z}}^{\mathbf{s}}\right), \\
& \mathbf{q}^{\mathbf{s}}=\mathbf{Q} \cdot \hat{\mathbf{z}}^{\mathbf{s}}, \  
\mathbf{k}^{\mathbf{s}}=\mathbf{K} \cdot\left(\hat{\mathbf{z}}^{\mathbf{s}} \oplus \hat{\mathbf{z}}^{\mathbf{d}}\right), \  
\mathbf{v}^{\mathbf{s}}=\mathbf{V} \cdot\left(\hat{\mathbf{z}}^{\mathbf{s}} \oplus \hat{\mathbf{z}}^{\mathbf{d}}\right),
\end{aligned}
\end{equation}
where $\hat{\mathbf{z}}^{\mathbf{d}}$ and $\hat{\mathbf{z}}^{\mathbf{s}}$ are latent depth and semantic embeddings within transformer blocks, $\oplus$ denotes concatenation, and $\mathbf{Q}$, $\mathbf{K}$, and $\mathbf{V}$ are the query, key, and value projection matrices. The final cross-task features are computed as:
\vspace{-3pt}
\begin{equation}\vspace{-3pt}
\boldsymbol{\operatorname { A t t }}\left(\mathbf{q}^i, \mathbf{k}^i, \mathbf{v}^i\right), \  i \in\{d, s\},
\end{equation}
where $\mathbf{Att}(\cdot)$ represents softmax attention.
\section{Experimental Details}
\label{supp:experiment}
\subsection{Category-level Pose Estimation.}
Following the evaluation protocol in~\cite{SwinDR}, we adopt the SwinDRNet backbone and introduce two decoder heads to predict the coordinates of the Normalized Object Coordinate Space (NOCS) map and the semantic segmentation mask. Pose fitting is then performed by aligning the reconstructed object point clouds in the world coordinate space with the predicted object point clouds in the normalized object coordinate space. Finally, we can estimate the 6D object pose through this alignment process.
\subsection{Robotic Grasping.}
For robotic grasping experiments, we use a UR5 robotic arm equipped with a Robotiq-85 end effector. RGB-D frames are captured using an Intel RealSense D415 camera. Some objects in our experiments feature non-Lambertian surface materials, such as glass, transparent plastic, and metal, which pose significant challenges for both depth sensing and robotic grasping. We first apply depth completion methods to correct inaccurate depth measurements. Then, leveraging the camera intrinsics, we reconstruct the corresponding point cloud, which is subsequently processed by GraspNet~\cite{GraspNet} to predict grasp poses.

\section{More Qualitative Results.}
\label{supp:qualitative}
\begin{figure*}[!htbp]
	\centering
    \vspace{2em}
	\includegraphics[width=1\linewidth]{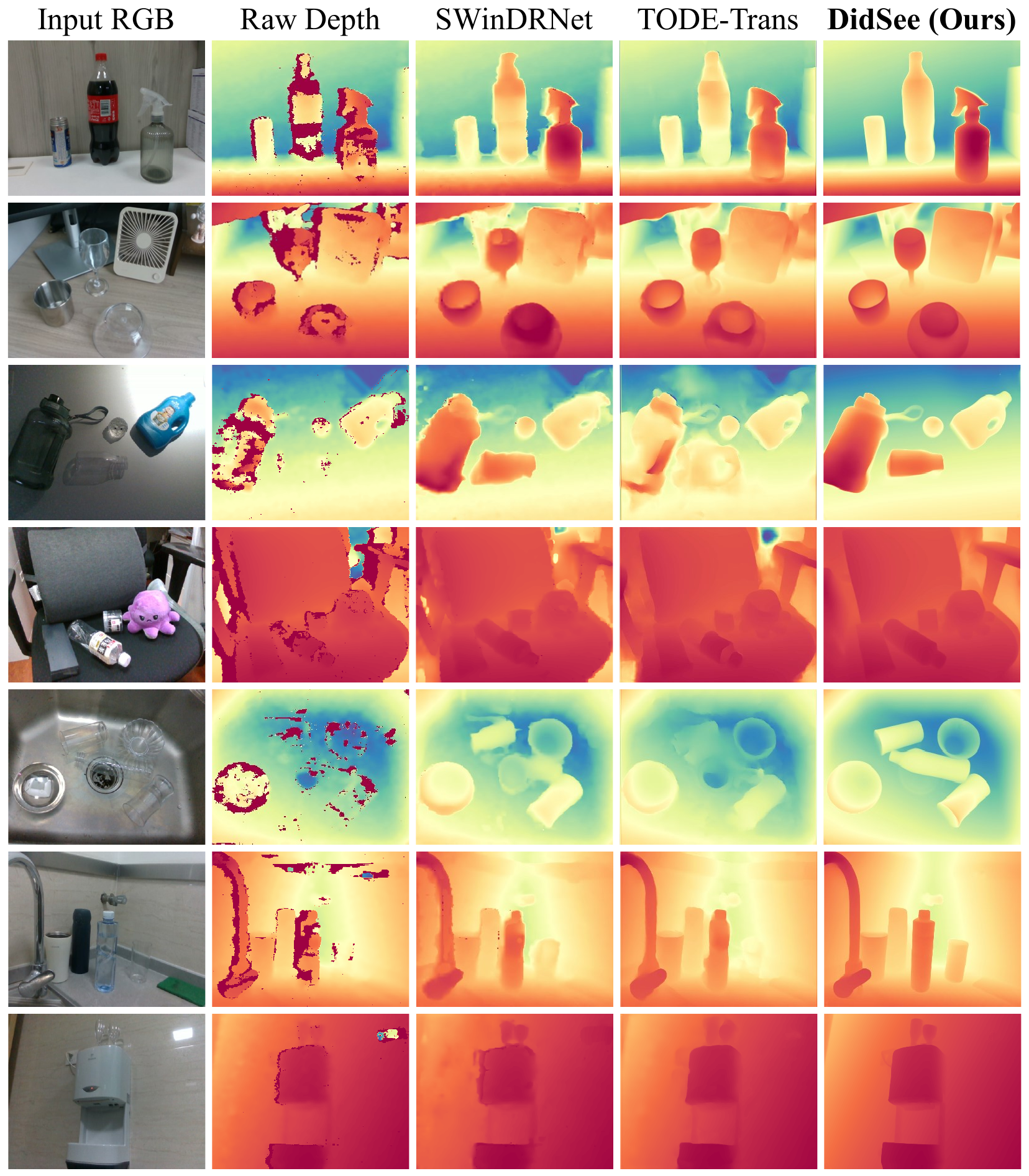}
	\caption{\textbf{Qualitative comparison on in-the-wild scenes (1/2).}}
    \vspace{2em}
	\label{fig:supp_figs1}
\end{figure*}

\begin{figure*}[!htbp]
	\centering
    \vspace{2em}
	\includegraphics[width=1\linewidth]{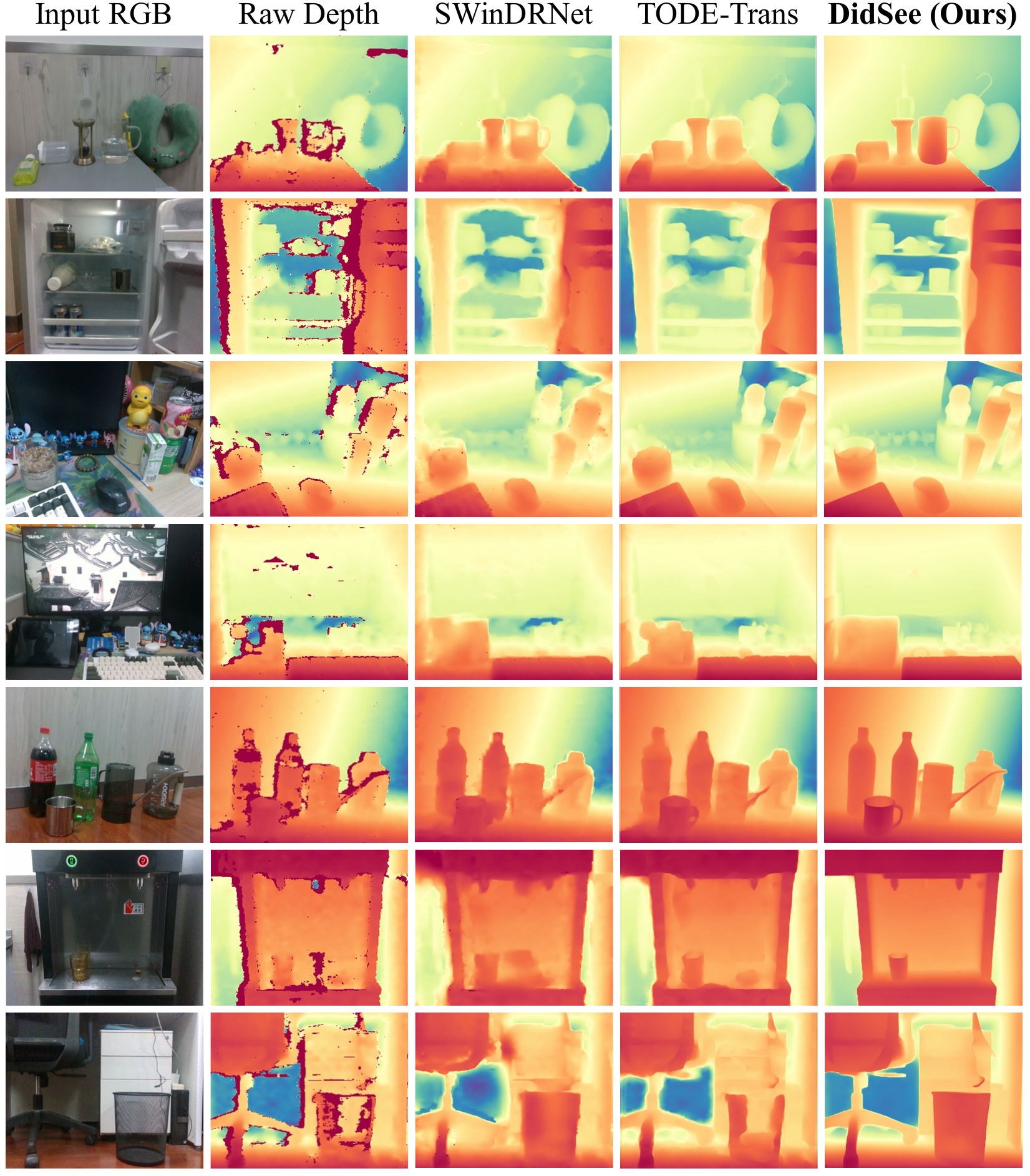}
	\caption{\textbf{Qualitative comparison on in-the-wild scenes (2/2).}}
    \vspace{2em}
	\label{fig:supp_figs2}
\end{figure*}

We present qualitative comparisons with state-of-the-art methods~\cite{TODE-Trans,SwinDR} across diverse in-the-wild scenes in \cref{fig:supp_figs1} and \cref{fig:supp_figs2}. The scenes contain objects with varied materials under challenging conditions, such as non-planar layouts, heavy occlusion, and low-light environments. Despite being trained solely on the synthetic tabletop dataset DREDS-CatKnown~\cite{SwinDR}, DidSee generates highly accurate and fine-grained depth maps, particularly in non-Lambertian regions. This highlights the strong generalizability of DidSee to real-world scenarios.

\section{Limitations}
\label{supp:limitation}
While DidSee has demonstrated remarkable performance, its reliance on Stable Diffusion~\cite{Stable-Diffusion}, which contains a substantial number of parameters, presents a challenge in terms of efficiency. In our future work, we will delve deep into a more flexible version of DidSee to enhance efficiency and reduce computational overhead.

